\documentclass[conference]{IEEEtran}
\IEEEoverridecommandlockouts
\usepackage{cite}
\usepackage{amsmath,amssymb,amsfonts}
\usepackage{algorithmic}
\usepackage{graphicx}
\usepackage{textcomp}
\usepackage{booktabs}
\usepackage{xcolor}
\usepackage{subcaption}
\def\BibTeX{{\rm B\kern-.05em{\sc i\kern-.025em b}\kern-.08em
    T\kern-.1667em\lower.7ex\hbox{E}\kern-.125emX}}
\begin{document}

\title{Prediction is Better than Detection: Traffic Congestion Control using Drones\\
{
\thanks{This work was conducted within the 3D-Net project (project no. 923328), funded under the RTI Mobility Initiative by the Austrian Federal Ministry for Innovation, Mobility and Infrastructure (BMIMI) and the Austrian Research Promotion Agency (FFG). The project is also part of the CELTIC-NEXT Eureka Cluster (project ID C2025/1-11).}
}
}

\author{\IEEEauthorblockN{1\textsuperscript{st} Samira Hayat}
\IEEEauthorblockA{\textit{Lakeside Labs GmbH} \\
Klagenfurt, Austria \\
0000-0003-1725-4106}
\and
\IEEEauthorblockN{2\textsuperscript{nd} Christian Raffelsberger}
\IEEEauthorblockA{\textit{Lakeside Labs GmbH} \\
Klagenfurt, Austria \\
0000-0002-8366-144X}
}

\maketitle

\begin{abstract}
A central question in deploying teams of mobile robots for persistent monitoring is how task performance scales with fleet size, and whether this scaling holds once sensing drives downstream action rather than mere observation. We study this question for a team of drones performing traffic-jam detection and prediction in a simulated road network, whose reports drive an adaptive traffic-signal controller in closed loop. We build a multi-agent simulation, with vehicles following Nagel-Schreckenberg cellular-automaton dynamics and drones patrolling junctions via a round-robin policy, and sweep fleet size, traffic level, and network size to evaluate detection rate, detection delay, and prediction rate. We show how performance plateaus for fleet size approximating the number of junctions being monitored, and offer a general fleet-provisioning rule for persistent-monitoring deployments. More significantly, adapting the signal on a predicted jam, rather than a detected one, roughly doubles the resulting reduction in jam duration, showing that the value of onboard prediction in a sensing-to-action pipeline can exceed the value of adding more robots. Prediction accuracy, not sensing coverage, is now the binding constraint on further improvement, pointing to onboard inference, not fleet size, as the more promising direction for future work.
\end{abstract}

\begin{IEEEkeywords}
Drone fleet, UAVs, Intelligent transport system, Signal control, Jam detection, Jam prediction.
\end{IEEEkeywords}

\section{Introduction}

Persistent monitoring with mobile robot teams is increasingly used not just to observe an environment but to act on it: inspecting, alerting, or, as in this paper, directly controlling infrastructure in response to what is sensed. This closes the loop: a robot's observation triggers a decision that changes the environment, which the robot then re-observes on its next pass. This raises two questions distinct from pure sensing performance: does fleet-size scaling look the same once sensing feeds a downstream controller as it does for sensing alone, and is it more valuable to add robots or to make the sensing smarter, i.e., move from reactive sensing to sensing that predicts?

We study this question with a team of drones, or unmanned aerial vehicles (UAVs) performing traffic-jam detection and prediction over an urban road network, whose reports drive an adaptive signal controller in closed loop. Traffic congestion is a natural testbed: it is distributed and time-varying, fixed sensing infrastructure is costly to deploy densely, and UAVs can reposition to observe wherever congestion forms~\cite{b1,b2,b3}. 
The existing body of work treats the UAVs purely as a sensing system, without closing the loop to a control action or characterizing how closed-loop performance scales with fleet size. The one work that does close this loop, to our knowledge, acts reactively: a UAV-camera-fed deep reinforcement learning controller has been shown to restore already-congested networks to free flow~\cite{b11}, but only after congestion has fully manifested, and without characterizing its benefit as a function of fleet size. This paper addresses both gaps.

Our contributions are:

\begin{itemize}
    \item A multi-agent simulation testbed, built on the Mesa framework, coupling cellular-automaton traffic dynamics, signal-controller agents, patrolling UAVs, and stochastic disturbances, enabling controlled closed-loop experiments.
    \item A precursor-based jam detection and early-prediction criterion, calibrated to the fundamental diagram of the underlying traffic model.
    \item A controlled sweep of fleet size, traffic level, and road network size, showing their joint impact on detection and prediction performance, resulting in a practical fleet-sizing guideline for a given network topology.
    \item Evidence that triggering control on a predicted, rather than a detected event, brings more benefit.
\end{itemize}

The remainder of this paper is organized as follows. Sec.~II positions this work relative to prior UAV-based monitoring and adaptive signal control research. Sec.~III describes our simulation environment and jam detection/prediction criteria. Sec.~IV details our experimental setup. Sec.~V presents and discusses our results and Sec.~VI concludes the paper. 

\section{Related Work}

\subsection{UAV-Based Traffic Congestion Detection and Prediction}

UAVs have been used extensively for traffic monitoring. The pNEUMA experiment, flying ten drones over roughly 100 intersections, remains the largest empirical demonstration of aerial traffic sensing \cite{barmpounakis2020pneuma}, and drone/loop-detector fusion has been shown to improve flow forecasting under sparse ground sensing \cite{xiong2025multisource}. On the prediction side, precursor-based congestion detection is well studied on the ground: early bottleneck's growth predicts eventual jam severity \cite{duan2023spatiotemporal}, and transitional-instability indicators flag highway breakdown before it fully forms \cite{tak2026early}. However, neither evaluates such precursor signals for signal adaptation and jam alleviation. We address this gap by using prediction to reduce jam duration through adaptive signal control.

\subsection{Adaptive and Sensor-Triggered Signal Control}

A parallel body of work adjusts signal timing in real time based on sensed congestion conditions.  \cite{diakaki2000multivariable} proposes a traffic-responsive urban control (TUC) strategy, using a multivariable regulator approach to establish coordinated, network-wide responsive control suitable for saturated conditions. \cite{feng2015realtime} uses connected-vehicle sensing to enable fully real-time adaptive control without fixed detector infrastructure. More recent work targets queue spillback: a decentralized, communication-free controller that maximizes network throughput under spillback using shockwave-based queue estimation~\cite{noaeen2021realtime}, a predicted-demand controller that provides globally optimal signal timing for isolated oversaturated intersections~\cite{mohajerpoor2023optimal}, and an event-based controller that classifies and responds to turning-bay overflow and mutual queue blockage on congested arterials~\cite{lin2025realtime}. Camera-based feedback uses vision-sensed flow as the model-predictive control input~\cite{park2024traffic}. Composite congestion indicators elsewhere combine at most two of density, speed, or flow~\cite{trinh2021fuzzy}, and precursor-based prediction has instead relied on critical-slowing-down theory~\cite{chattopadhyay2026multivariate} rather than a relaxed detection threshold; by contrast, our detector and predictor share a single novel three-variable (density, speed, queue) basis, differing only in threshold strictness.

Most closely related to this work, Guo et al. propose AVARS, a UAV-camera-fed deep reinforcement learning controller that reacts to already-manifested congestion and is shown to restore free-flow conditions within a typical UAV battery life~\cite{b11}. AVARS establishes that aerial sensing can directly drive adaptive signal control, but purely reactively. This work extends that direction along two axes: our controller is triggered by both detection \emph{and} prediction reports rather than detection alone, and we characterize how detection and prediction rate, and downstream control benefit jointly scale with UAV fleet size, network size, and traffic level. 

\section{System Model}
\subsection{Traffic Modeling}
\label{sec:simenv}
We implement a discrete-time, agent-based urban traffic simulator using the Mesa framework~\footnote{https://github.com/samirahayat/3D-NET-JamPrediction.git}. The environment consists of four interacting components: (1) \textit{traffic agents}, vehicles moving through a signalised road network with four-way intersections; (2) \textit{infrastructure agents}, comprising a ground station (GS), junction controllers and signalized approach lights; (3) \textit{aerial sensing agents}, UAVs that patrol junctions to monitor congestion; and (4) \textit{exogenous disturbances}, stochastic road events such as accidents or temporary lane blockage. No additional ground-based sensors are used to monitor traffic or detect jams.

\begin{figure}[!t]
    \centering
    \includegraphics[width=0.8\linewidth]{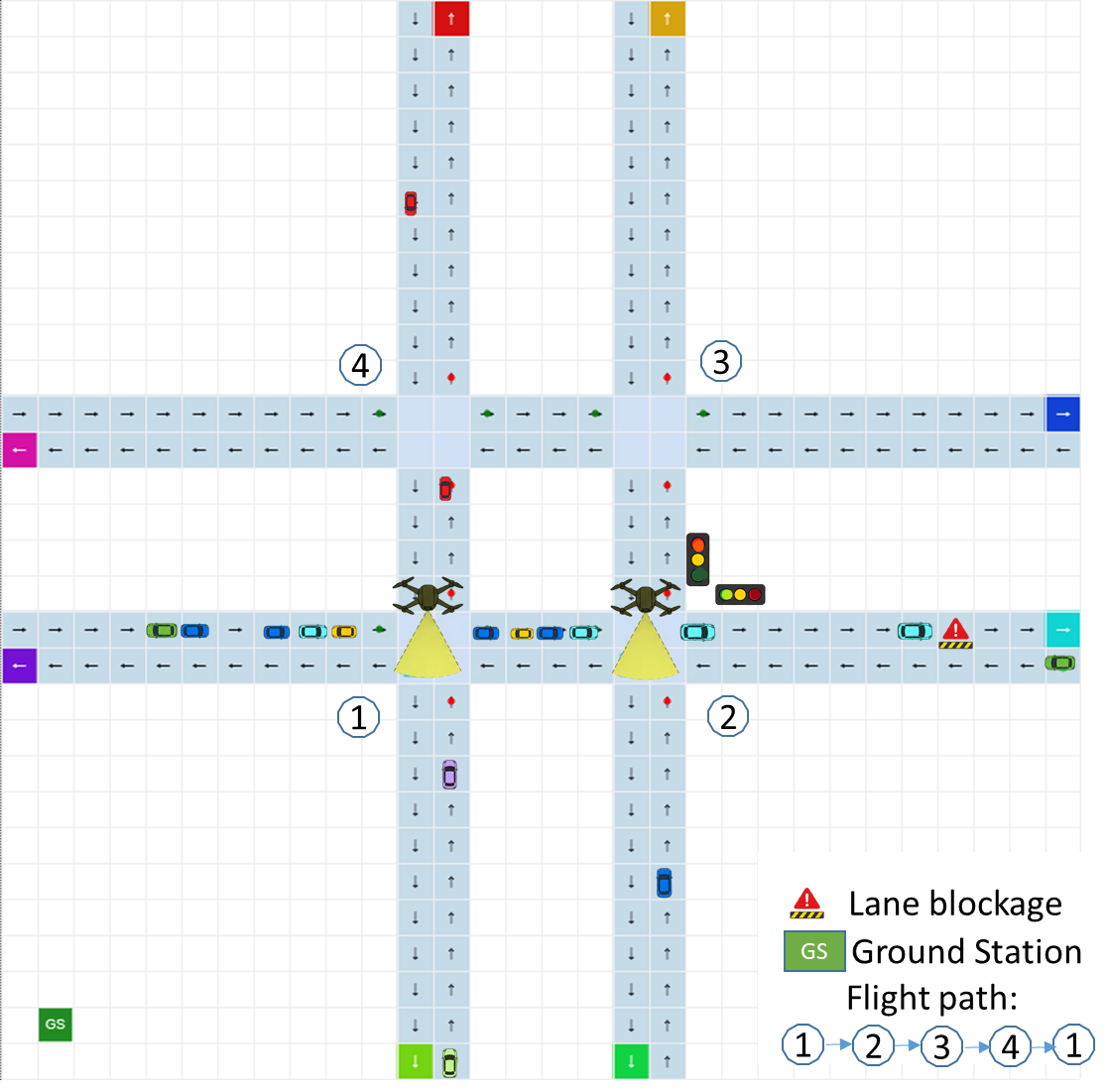}
    \caption{Traffic environment. The different car colors indicate the intended exit cell. All junctions are signalised, similar to upper-right corner of junction 2.}
    \label{fig:6Diagram}
\end{figure}

Road intersections are controlled by traffic signals consisting of alternating horizontal and vertical green phases separated by yellow transitions (see Fig.~\ref{fig:6Diagram}). Vehicles are introduced through probabilistic spawning at network entry points. Each road provides one entry cell per direction of travel, so the total number of entry cells scales with the number of roads in the topology. At each tick, three independent spawn attempts are made at randomly selected entry cells, yielding an expected inflow $E[N_{\text{spawn}}] = 3p$ vehicles/tick, where $p$ is the per-attempt spawn probability.

Vehicles travel on directed lane segments governed by the Nagel-Schreckenberg (NaSch) cellular automaton model \cite{NagelSch1992} with maximum speed $v_{\max}=4$ cells/tick and stochastic deceleration probability $p=0.3$. Following the original model's approach, each cell length is $7.5$ m (average space occupied by a vehicle, including spacing to other vehicles), and each simulation tick corresponds to 1 s. At each tick, the simulator updates stochastic obstacles, followed by vehicle injection and traffic evolution (subject to collision avoidance and signal constraints); ground truth congestion events and UAV detection and prediction outcomes are then updated.

With $v_{\max}=4$ cells/tick and $p=0.3$, the corresponding NaSch fundamental diagram exhibits a free-flow region for $\rho \lesssim 0.15$, a transitional regime near capacity for $0.15 \leq \rho \leq 0.25$, and a congested branch for $\rho \gtrsim 0.25$. The free-flow mean speed comes out to be $3.0$--$3.5$ cells/tick. These values serve as the baseline for the choice of congestion thresholds in the following section.

\subsection{Jam Modeling}
\label{sec:JamFormulation}

We employ a multi-criteria trigger to classify a junction approach direction as ``jammed'', requiring three conditions to hold simultaneously during the pure-green signal phase. Let $\rho$ denote the vehicle density in a 15-cell window upstream of the stop line, $\bar{v}$ the mean vehicle speed within that window, and $Q$ the contiguous queue length at the stop line. The jam $J$ on an approach is conditioned upon all three:
\begin{equation}
J=
\begin{cases}
1,&
\rho\ge\rho_{\min}
\land
\bar v<\beta v_{\max}
\land
Q\ge Q_{\min},\\
0,&\text{otherwise},
\end{cases}
\end{equation}
where $\rho_{\min} = 0.30$, $\beta = 0.40$ (yielding a speed threshold of $1.6$ cells/tick), and $Q_{\min} = 1$.

This composite metric addresses the ambiguity inherent in flow-based indicators ($q = \rho \cdot \bar{v}$), which cannot distinguish between an empty road and a fully gridlocked one, both of which exhibit zero flow. The threshold values are calibrated against the NaSch fundamental diagram to ensure robustness:
\begin{itemize}
    \item \textbf{Density Gate ($\rho_{\min} = 0.30$):} Set above the critical density ($\rho \approx 0.25$); a lower value would cause false positives near capacity.
    \item \textbf{Speed Gate ($\beta = 0.40$):} The threshold of $1.6$ cells/tick is approximately half the free-flow mean speed ($3.0$-$3.5$ cells/tick), ensuring that the criterion triggers only on genuine gridlock.
    \item \textbf{Queue Gate ($Q_{\min} = 1$):} Requires at least one vehicle queued at the stop line; combined with high density and low speed, this confirms the slowdown has propagated to the intersection. Larger queue thresholds suppress short but operationally significant congestion events, particularly obstacle-induced congestion.
\end{itemize}

A jam event opens only after the trigger condition persists for $3$ consecutive pure-green ticks. During following red phases, an active jam remains open. 

\subsection{UAV Monitoring Architecture}

UAVs patrol junctions using a round-robin route augmented with lightweight conflict avoidance: if a UAV's next assigned junction is currently monitored by another drone, it proceeds to the next unoccupied junction along its route. This avoids wasting patrol time at an already-monitored junction without requiring global path optimization. This keeps the policy substantially simpler than adaptive or optimized patrol strategies, allowing fleet size to remain the primary driver of monitoring performance in our experiments. Each UAV travels at 2 cells/tick (15 m/s) and performs no sensing while in transit. Upon reaching a junction, it hovers for
\begin{equation}
t_{\mathrm{hover}}=t_{\mathrm{proc}}+t_{\mathrm{cycle}},
\end{equation}
where $t_{\mathrm{proc}}\sim\mathcal{U}(10,40)$ ticks represents sensor-processing latency, informed by realistic onboard sensor specifications, while $t_{\mathrm{cycle}}$ corresponds to one complete signal cycle (horizontal green/yellow plus vertical green/yellow). No detection is registered during $t_{\mathrm{proc}}$. After $t_{\mathrm{proc}}$ elapses, the UAV performs active scanning of all incoming approaches for $t_{\mathrm{cycle}}$, ensuring both horizontal and vertical approaches are observed during their green phases; a shorter visits could miss the relevant right-of-way phase, since jam detection is signal-phase-gated.

\subsubsection{Detection Logic}
During active scanning, the UAV applies the same three-criterion jam detector used for ground truth. As with ground truth, a direction is reported as jammed only if all three trigger conditions are satisfied; failure of any criterion resets the jam state to false.

\subsubsection{Prediction Logic}
For approaches not currently classified as jammed, the drone generates an early congestion prediction using the temporal density gradient $\Delta k$, computed as the difference between the current observed density $\rho(t)$ and the last density observation stored at the ground station for that specific approach, $\rho_{\text{GS}}(t_{\text{prev}})$:
\begin{equation}
\Delta k = \rho(t) - \rho_{\text{GS}}(t_{\text{prev}})
\end{equation}

This approach separates sensing from historical state maintenance: UAVs provide current observations, while the ground station maintains the information required to identify trends.

Intuitively, a jam is predicted when an approach is nearing the congestion threshold and a queue has begun forming at the stop line, together with either a marked slowdown in traffic speed or a worsening density trend:
\begin{equation}
\begin{split}
    P=
\begin{cases}
1,&
\rho\ge\alpha\rho_{\min}
\land
(\bar v< 1.6\beta v_{\max} \lor \Delta k>\gamma)
                      \land Q\ge 1,\\
0,&\text{otherwise},
\end{cases}
\end{split}
\end{equation}
where $\alpha<1$ scales the congestion density threshold. In our case, the density gate is relaxed to $60\%$ of the jam detection threshold ($\alpha = 0.6$) to capture early loading. The gradient threshold $\gamma = 0.02$ corresponds to approximately $0.3$ additional vehicles in the 15-cell window, a magnitude that exceeds typical tick-to-tick noise but indicates a genuine worsening trend. Predictions are validated against ground truth over a horizon of $H_p$ ticks; if a jam initiates within this window, the prediction is marked correct, and the lead time is calculated as the interval between the prediction timestamp and the jam onset. 

Each drone sends its traffic report to the ground station upon completing its hover task. The traffic report contains the traffic density at each approach, any detected jam and its detection time, and any predicted jam and its prediction time.

\subsection{Adaptive Signal Control}
Upon receiving the traffic report from a UAV, the ground station signals the junction controller to adapt its signal timing instantaneously to mitigate the developing bottleneck. The control logic operates on a base green duration $G_0$ ticks and adjusts it based on the congestion status of the current and opposing approach axes.

Let $C_H$ and $C_V$ be binary indicators denoting whether the horizontal (E-W) and vertical (N-S) axes are congested, respectively. Congestion is determined by the ground station using macroscopic criteria analogous to the jam modeling. The controller computes a green extension $\delta G \in [-G_0/2, +G_0/2]$ ticks according to the following rules:
\begin{itemize}
    \item If only the current axis is congested, $\delta G = +G_0/2$.
    \item If only the opposing axis is congested, $\delta G = -G_0/2$.
    \item If both axes are congested, priority rules determine extension, defaulting to no change if priorities are equal.
    \item If no congestion is detected, $\delta G = 0$.
\end{itemize}
The effective green duration is $G_{\mathrm{eff}} = \max(G_{\text{min}}, G_0 + \delta G)$, where $G_{\text{min}}$ is the minimum green signal length. A pre-emptive shortening mechanism is triggered if a prediction indicates high incoming flow ($> 0.3$ vehicles/tick) on the opposing axis before it becomes critically congested, reducing the current green phase by an additional $G_0/4$ ticks to yield right-of-way proactively. All adaptations reset at the end of each green phase to ensure responsiveness to real-time conditions.

\subsection{Exogenous Disturbances}
To simulate non-recurrent disruptions such as accidents or roadworks, we introduce stochastic obstacles. At each tick, an obstacle spawns with probability $0.04$ if fewer than two obstacles are currently active. Obstacles persist for a uniform random duration between $2G_{\text{min}}$ and $2G_0$ ticks. The probability and duration ranges were selected to represent typical minor incident frequencies in urban grids.

\section{Experimental Setup}

We perform simulations over a grid of $120 \times 120$ cells, varying three parameters to characterize the performance of the UAV monitoring system.


\noindent \textbf{Traffic Level:} $p$ is the probability that an individual vehicle-spawn attempt succeeds. With three attempts per tick, this yields an expected inflow $E[N_{\text{spawn}}]=3p$ vehicles/tick. We evaluate spawn probabilities $p = 0.25, 0.5, 0.75$, referred to as low, medium, and high traffic levels, respectively, intended to characterize low traffic through the saturation boundary.

\noindent \textbf{Drone Fleet Size:} $N_u$ is varied from 1 to 10 drones, spanning sparse monitoring, where junctions can go unobserved for extended periods, to dense deployments with diminishing monitoring returns.

\noindent \textbf{Network Size:} We vary the number of junctions from 2 to 10 in the simulation grid. Each network size corresponds to a distinct grid topology, parameterized by its number of horizontal ($H$) and vertical ($V$) roads, with junction count equal to $H \times V$: the 2-junction network uses $H=1, V=2$; the 6-junction network uses $H=2, V=3$; and the 10-junction network uses $H=2, V=5$. With one NaSch cell~$=7.5$ m and a $120 \times 120$ grid, the simulated area is $0.81\,\text{km}^2$, so the three network sizes correspond to junction densities of $2.47$, $7.41$, and $12.35$ junctions/km$^2$, respectively. The two lower densities fall well below the built-up-area threshold reported in \cite{Zho18}, while the highest (12.35 junctions/$\text{km}^2$) exceeds the lowest built-up threshold (10 junctions/km², for the smallest town surveyed). Nonetheless, the strict 4-way grid topology imposed at each junction in our case produces traffic dynamics distinct from an unsignalized road network of comparable sparsity: as the number of junctions increases from 2 to 10 within the same 0.81 $\text{km}^2$ area, the dominant traffic regime shifts from a largely uninterrupted, free-flow arterial corridor to an interrupted-flow network shaped by signalling and availability of alternative routes. 

All other simulation parameters are held constant across runs to ensure comparability. The reported results are a mean of 20 seeded simulations, run for $2000$ ticks per run, sufficient for multiple jam formation and clearance cycles to achieve statistical stability. The green signal duration $G_0 = 12$ and $G_{\text{min}}$ and yellow duration $= 4$ ticks, representative of standard signal plans for the network sizes considered. The prediction horizon $H_p = 12$ ticks is chosen equal to the length of the green signal, to balance early-warning utility against the risk of false alarms due to short-term volatility. These fixed values provide consistency across the experiments; their joint optimization, and matching traffic and disturbance parameters to real-world dataset, is outside the scope of this work.


\section{Results and Evaluation}

\begin{figure}[!t]
    \centering
    \begin{minipage}{0.5\textwidth}
        \centering
        \includegraphics[width=\textwidth]{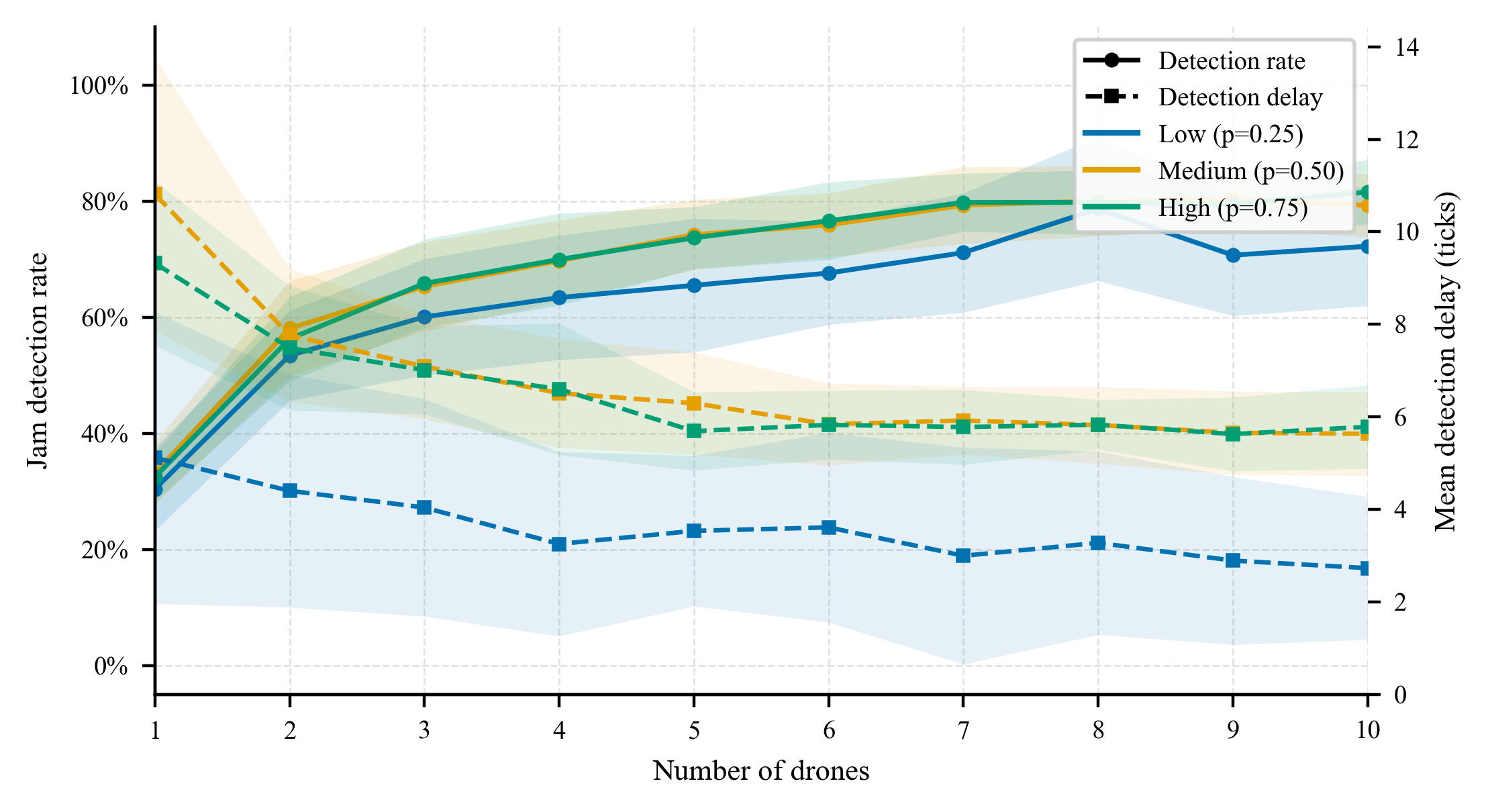}
        \subcaption{2 junctions}
        \label{fig:rate_2junc}
    \end{minipage}
    \begin{minipage}{0.5\textwidth}
        \centering
        \includegraphics[width=\textwidth]{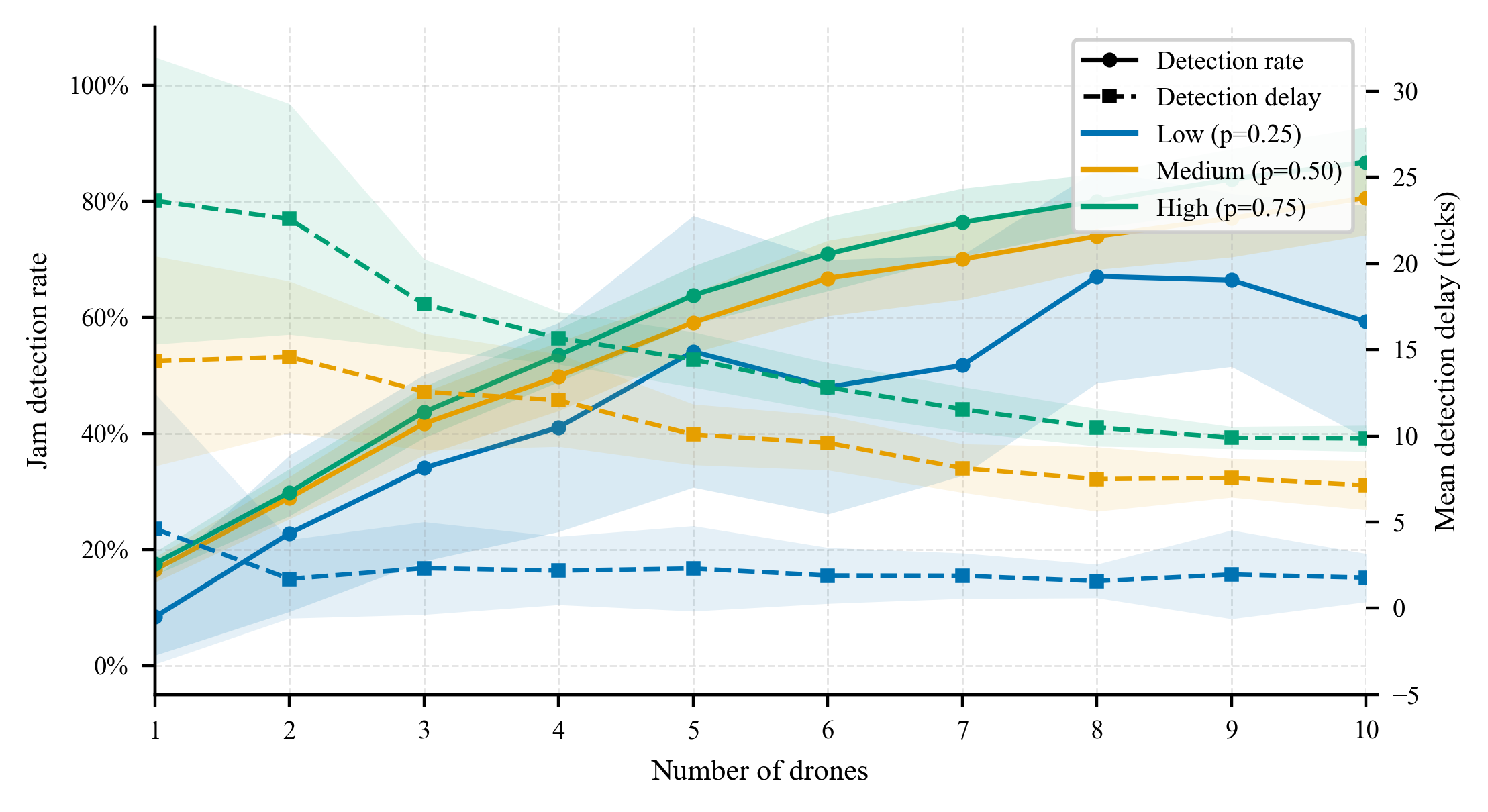}
        \subcaption{6 junctions}
        \label{fig:rate_4junc}
    \end{minipage}
    \begin{minipage}{0.5\textwidth}
        \centering
        \includegraphics[width=\textwidth]{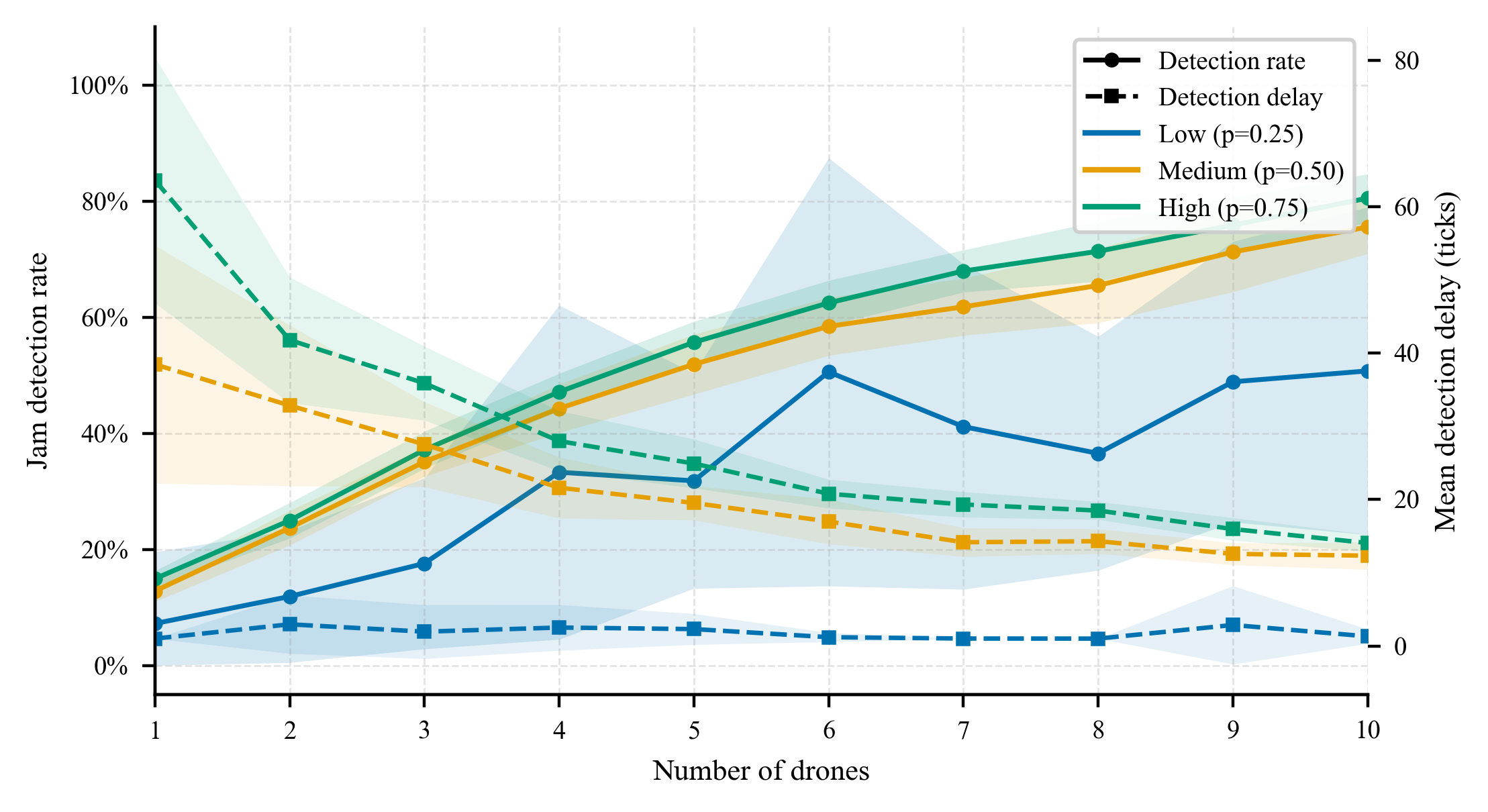}
        \subcaption{10 junctions}
        \label{fig:rate_10junc}
    \end{minipage}\hfill
    \caption{Jam detection rate and delay versus fleet size.}
    \label{fig:Detection_rate_delay}
\end{figure}





\subsection{Detection Rate and Detection Delay}

We examine how detection rate and detection delay are affected by fleet size, traffic level, and network size. Figs.~\ref{fig:rate_2junc}, ~\ref{fig:rate_4junc} and ~\ref{fig:rate_10junc} show jam detection rate and detection delay (mean and standard deviation) as a function of the number of drones, for the three traffic levels ($p = {0.25, 0.5, 0.75}$) and the three network sizes (2, 6, and 10 junctions). The solid lines represent the detection rate, while dashed lines show detection delay.

As the number of drones increases, detection rate rises and detection delay falls across all three network sizes. This improvement plateaus for fleet size approximately equal to the number of junctions: adding more drones beyond this point still yields modest additional gains in detection rate, but these gains quickly taper off. Because drones patrol in a round-robin fashion -- hovering at each junction for $t_{\text{hover}}$ before moving to the next -- junctions are not continuously covered even when fleet size equals the number of junctions, since travel time between junctions necessarily leaves gaps in coverage. A comparable improvement beyond this plateau could likely be obtained at a fixed fleet size equal to the number of junctions through a more sophisticated patrol strategy (e.g., adaptive scheduling) rather than by adding more drones.

\begin{figure*}[!t]
    \centering
    \includegraphics[width=0.94\textwidth, trim={0 1cm 0 0.9cm},clip]{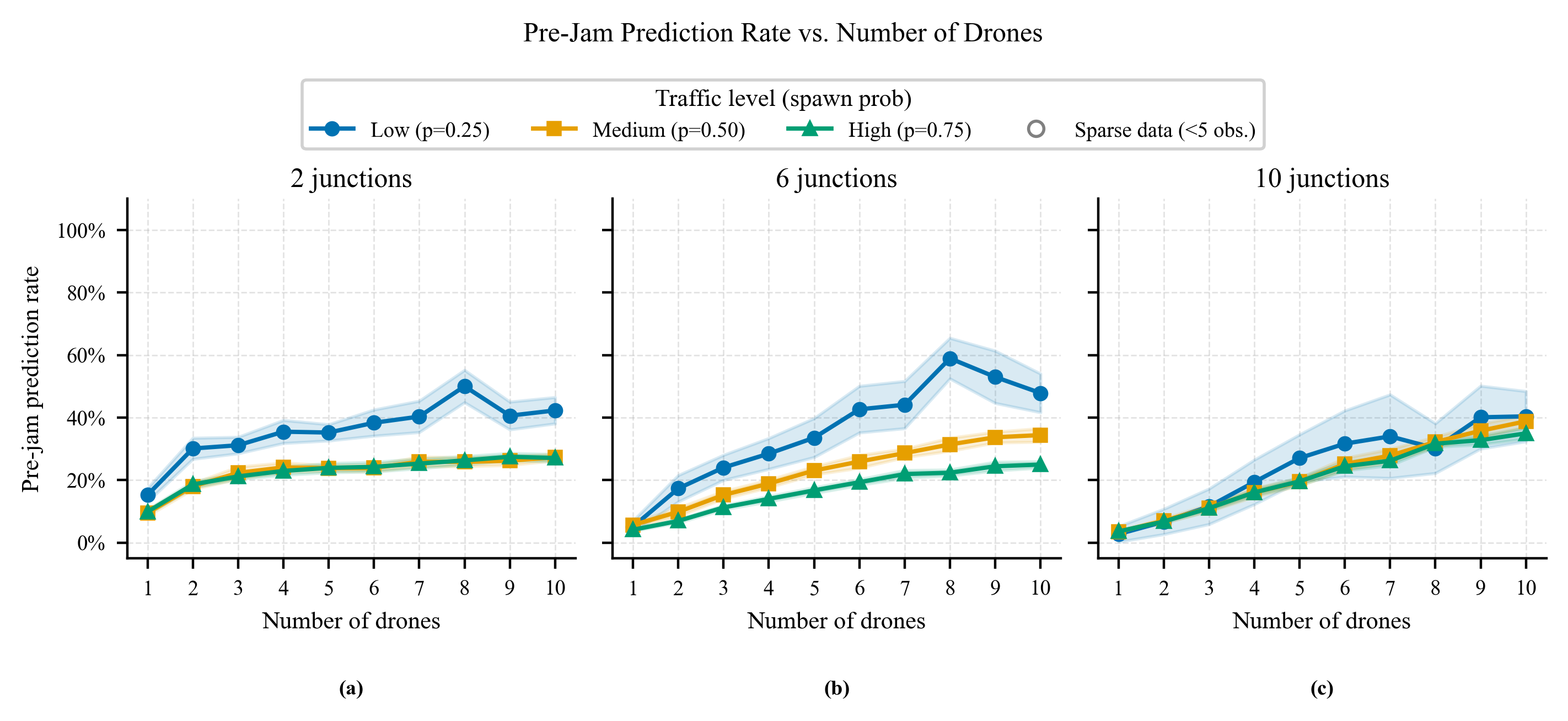}
    \caption{Pre-jam prediction rate versus fleet size}
    \label{fig:predrate}
\end{figure*}

We adopt round-robin patrolling for two reasons. First, its simplicity isolates the effect of fleet size from that of routing strategy. Second, it is robust to drones dropping out of service mid-operation -- for example due to battery depletion or failure -- since the remaining drones simply continue the shared route without coordination or knowledge of the current fleet size.

The plots also show that we never achieve a 100\% detection rate, even for the smallest network size (2 junctions; Fig.~\ref{fig:rate_2junc}). This is a consequence of our sensing parameters: sampling $t_{\text{proc}}$ uniformly between 10 and 40 ticks (reflecting the processing latency of real onboard sensors) introduces a delay before a drone can register a jam, during which the jam may already have cleared.

Detection delay also improves as fleet size increases. At low traffic level ($p=0.25$), drones detect jams almost immediately after they begin forming, regardless of network size. For the smallest network (2 junctions), medium and high level ($p = 0.5,0.75$) produce similarly low detection delays (Fig.~\ref{fig:rate_2junc}). This similarity breaks down at 6 junctions (Fig.~\ref{fig:rate_4junc}): the gap between medium and high level traffic detection delay is approximately 9.5 ticks with a single patrolling drone, narrowing to approximately 2.5 ticks with a fleet of 10. The same pattern appears at 10 junctions (Fig.~\ref{fig:rate_10junc}), but far more pronounced: with a single drone, the gap between medium and high traffic level delays reaches approximately 25 ticks, dropping sharply to approximately 9 ticks with just two drones, and converging to approximately 2.5 ticks at a fleet of 10, the same residual gap observed for 6 junctions. This one-to-two-drone drop is markedly steeper than any subsequent gain from adding drones, making the single-drone case a clear outlier.

We attribute this to the interval between consecutive visits to the same junction, which scales with the number of junctions and shrinks as fleet size grows. When the revisit interval is long relative to how quickly a jam forms -- as with a single drone patrolling 10 junctions -- there is more time for medium and high traffic jams to reach very different severities before they are detected, producing a large gap; This also explains why single-drone coverage is an outlier at 10 junctions: with 2 and 6 junctions, even one drone revisits often enough to keep the interval short. As fleet size grows, the revisit interval shortens toward the point where jams are detected shortly after forming regardless of traffic level, and the medium/high gap converges to a similar small residual across network sizes. This also explains why the 2-junction network shows little traffic-level-dependent gap even with a single drone: with only two junctions to cover, the revisit interval is already short at minimum fleet size.

\subsection{Prediction rate}

We now study the impact of increasing fleet size on jam prediction rate (mean and standard deviation) as network size increases from 2 to 10 junctions and traffic level increases from low to high (see Fig.~\ref{fig:predrate}). Using our chosen threshold values for prediction, the maximum prediction rate across our simulations does not exceed 55\%; nonetheless, the trends remain informative. As with detection rate, we observe a plateauing effect as fleet size increases. The highest prediction rates across all network sizes are achieved at low traffic level. For the 2-junction network, prediction rate is nearly identical for medium and high traffic levels, mirroring the pattern observed for detection rate. At 6 junctions, this pattern shifts: prediction rate is slightly higher for medium traffic than for high traffic -- the largest such gap observed across all three network sizes. At 10 junctions, the gap narrows again, with all three traffic levels producing similar prediction rates.

We attribute this pattern to two compounding factors. First, fleet size determines how frequently each junction is revisited, and because $t_{\text{proc}}$ -- the sensor setup delay drawn uniformly between 10 and 40 ticks -- must elapse at every visit before scanning begins, a drone patrolling more junctions wastes a larger cumulative share of its round-robin cycle on setup rather than active scanning. This is why fleet size must scale with junction count to reach the plateau observed for both detection and prediction: only once fleet size is comparable to junction count does each junction receive frequent-enough active-scan visits despite this fixed per-visit overhead. This revisit-driven mechanism also explains why low traffic level achieves the highest, most fleet-size-robust prediction rate at every network size: jams build up gradually enough under low demand that even a comparatively long revisit interval is still short enough to catch the precursor signal before the jam fully forms.

Second, and central to the medium/high traffic pattern across network sizes, our fixed traffic injection scheme distributes the same total traffic inflow over a larger road network as junction count grows, so each individual approach receives a smaller share of total demand in larger topologies. For the same nominal traffic level, the jams build up more gradually at each junction in the 10-junction network than in the 2-junction network, where the same total traffic concentrates on far fewer approaches. A more gradual build-up leaves a longer window for the density-gradient precursor signal to be observed before the jam fully forms, increasing prediction rate. This explains why the 2-junction network shows the lowest prediction rates for medium and high traffic: demand is concentrated enough that jams at both traffic levels form too abruptly for the precursor window to be reliably caught, regardless of fleet size. It also explains why the 10-junction network shows the highest prediction rate for medium and high traffic: demand is diluted enough that even high traffic jams build up gradually enough to be reliably caught, once fleet size is sufficient to overcome the revisit-interval cost described above. The 6-junction network sits between these two extremes: its degree of demand dilution is enough to slow medium-traffic jam formation to a pace the precursor signal can catch, but not enough to do the same for the faster-forming, high-traffic jams. This asymmetry -- one traffic level caught, the other not -- is why there is a gap between the prediction rates for medium and high traffic at 6 junctions, unlike at 2 and 10 junctions.

We note here that the fleet-sizing strategy raises detection and prediction reliability toward its achievable ceiling rather than raising that ceiling itself: our threshold-based prediction rule plateaus at a 55\% rate regardless of fleet size. Improving prediction accuracy further would require refining the detection criteria or precursor signal rather than adding drones.

\subsection{Signal Adaptation Benefit}

\begin{figure*}[!t]
    \centering
    \begin{minipage}{0.45\textwidth}
        \centering
        \includegraphics[width=\textwidth]{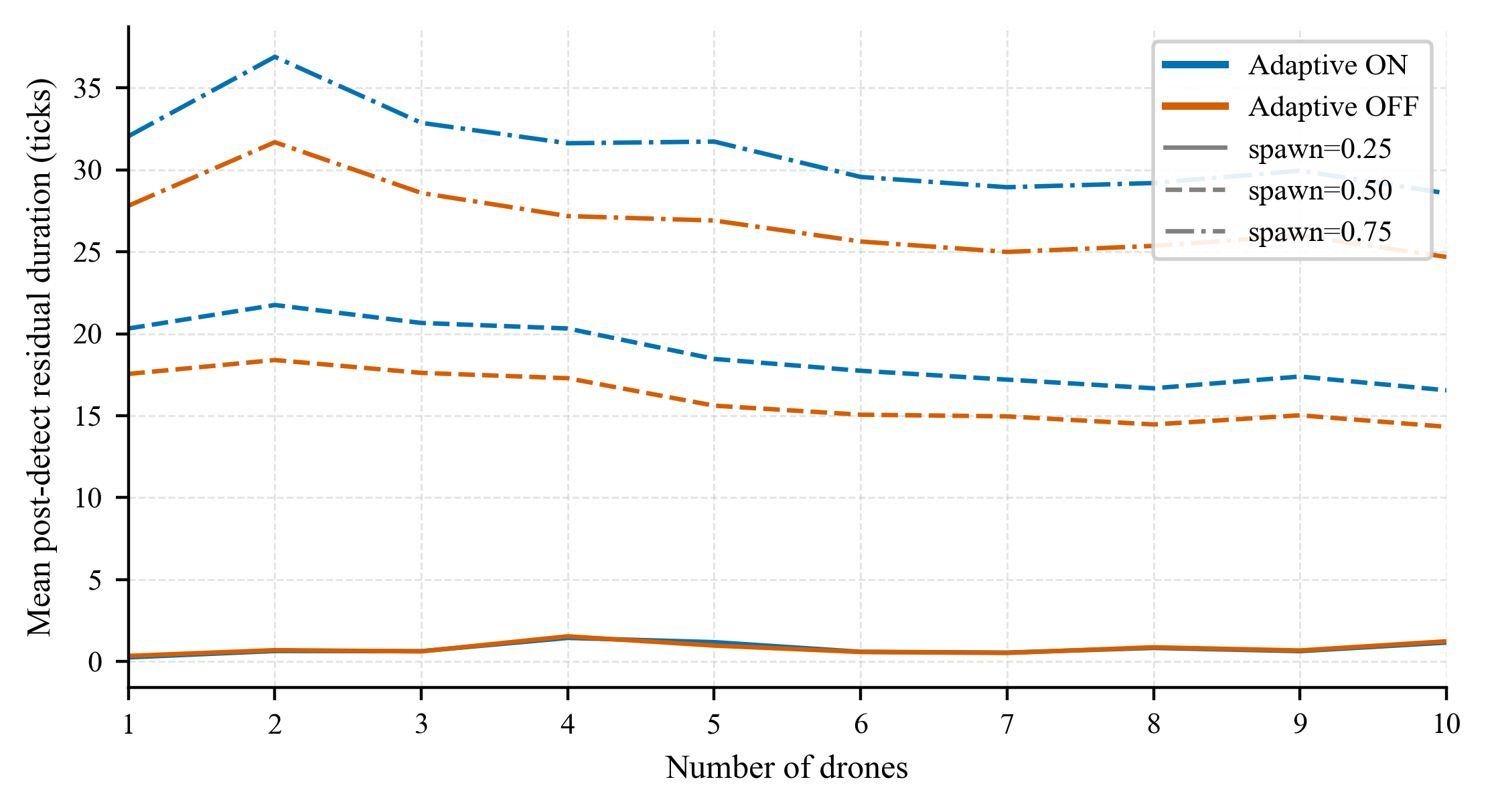}
        \subcaption{Detected versus undetected jams}
        \label{fig:detection}
    \end{minipage}
    \begin{minipage}{0.45\textwidth}
        \centering
        \includegraphics[width=\textwidth]{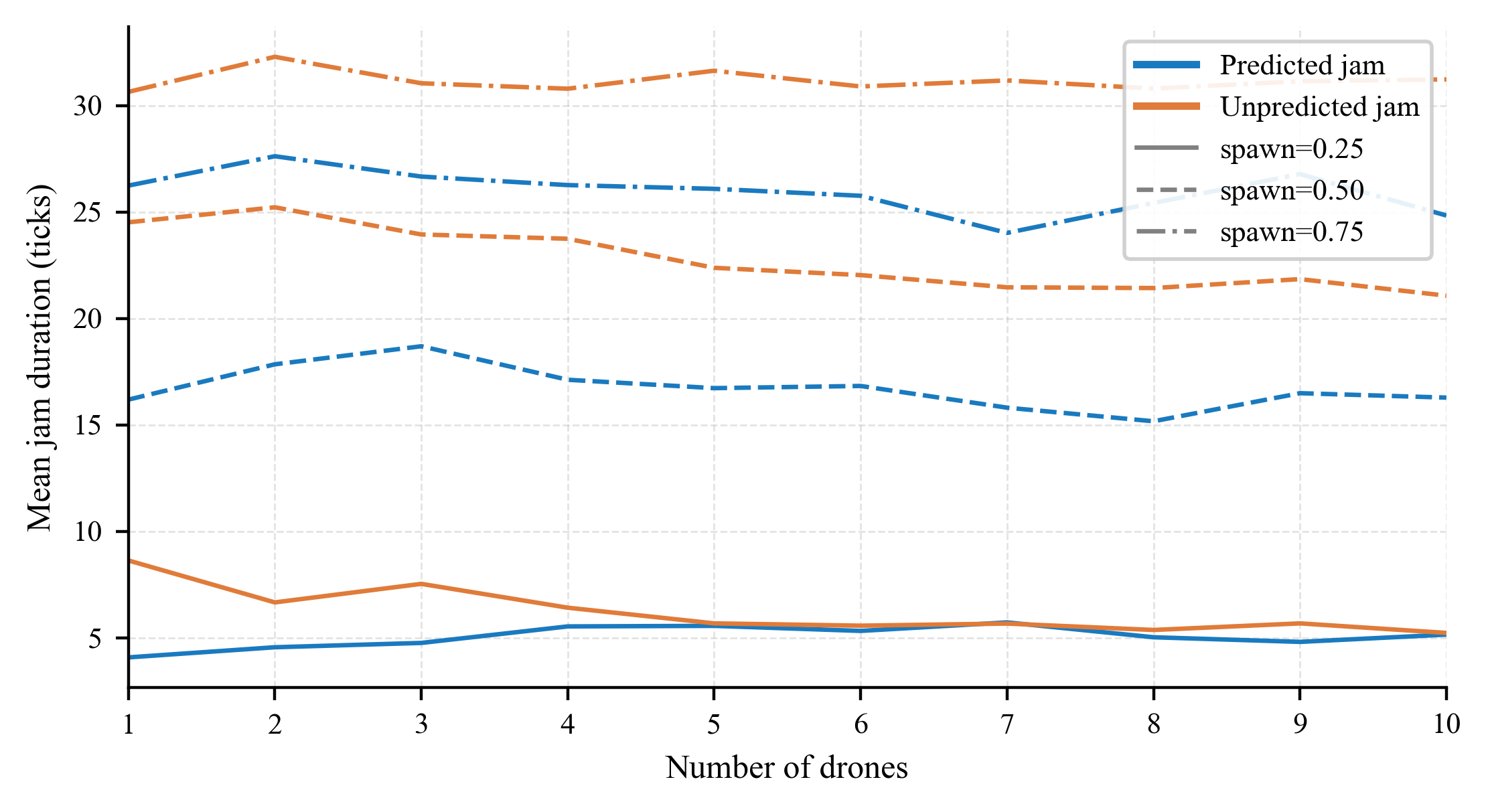}
        \subcaption{Predicted versus unpredicted jams}
        \label{fig:prediction}
    \end{minipage}\hfill
    \caption{Impact of signal adaptation on jam duration}
    \label{fig:benefit}
\end{figure*}

Finally, we report the overall improvement in jam duration achieved when signal control is adapted in response to detected and predicted jams, for a representative network size (6 junctions; Fig.~\ref{fig:benefit}). Fig.~\ref{fig:detection} shows the reduction in jam duration achieved when control is adapted following jam detection, relative to jams left unadapted. At low traffic level, adapting the signal yields no measurable improvement in jam duration. This improvement grows monotonically with traffic level, reaching a maximum of approximately 5 ticks at high traffic level, and this trend holds consistently regardless of fleet size.

Fig.~\ref{fig:prediction} shows the corresponding improvement in jam duration when a jam is predicted and the signal is adapted pre-emptively. Here we observe a substantially larger improvement in mean jam duration -- up to 8 ticks -- for jams that were correctly predicted, compared to jams that were not, across both medium and high traffic levels. In most cases, this prediction-driven improvement is roughly double the improvement achieved through detection alone.

We attribute the absence of benefit at low traffic level to a floor effect: at low demand, jams are infrequent and, when they do occur, tend to clear quickly even without adaptation, leaving little room for signal adjustment to shorten them further. As traffic level rises, unadapted jams persist longer -- queues rebuild faster than the signal can dissipate them -- giving the adaptive controller correspondingly more room to act, which explains the monotonic increase in benefit with traffic level. The independence of this benefit from fleet size follows from the mechanism itself: fleet size governs whether and how quickly a jam is detected or predicted in the first place, but once a jam has been correctly identified, the resulting signal adaptation is triggered identically regardless of which drone, or how many, made the identification. Finally, we attribute the larger benefit of prediction-triggered adaptation, roughly double that of detection-triggered adaptation in most cases, to the earlier point at which the controller intervenes: a predicted jam allows the pre-emptive reduction of $G_0/4$ to act on a queue that has not yet met our three-criterion jam threshold, while a detected jam can only be acted on after that threshold has already been reached, restricting the adaptation to mitigating an already-established queue rather than forestalling its growth. Intervening earlier in the congestion build-up therefore yields a proportionally larger reduction in eventual jam duration.

\section{Conclusion}

This paper presented a Mesa-based multi-agent simulation environment in which UAVs, calibrated against the fundamental diagram of the underlying Nagel–Schreckenberg traffic model, detect and predict junction-level congestion and trigger adaptive signal control accordingly. Our controlled sweep across fleet size, traffic level, and network size shows that detection and prediction performances improve sharply with fleet size only up to a network-size-dependent point -- approximately when fleet size matches junction count -- after which further drones yield diminishing returns. This acts as a practical provisioning guideline. Perhaps our most significant finding is that jam duration is reduced substantially more when signal control is triggered pre-emptively by a jam prediction than when triggered reactively after detection, roughly doubling the improvement achieved through detection alone in our experiments. This carries an implication beyond our specific system: since prior UAV-triggered signal-control work has been purely reactive, our results suggest that the predictive component of aerial sensing -- not merely its mobility or coverage -- may be the more consequential capability for future aerial congestion-management systems. At the same time, our threshold-based predictor plateaus well below a perfect prediction rate, indicating that the precursor signal itself, rather than sensing infrastructure, is now the primary bottleneck to further congestion pre-emption. Future work will explore richer precursor signals to raise the prediction-rate ceiling itself, and coordinated, adaptive patrol strategies that could offer better performance with fewer drones than the round-robin policy used here.


\vspace{12pt}

\end{document}